%% file: main.tex
\documentclass[11pt]{article}
\usepackage[margin=1in]{geometry}
\usepackage{graphicx}
\usepackage{booktabs}
\usepackage{longtable}
\usepackage{pdflscape}   
\usepackage{amsmath}
\usepackage{hyperref}
\usepackage{caption}
\usepackage{natbib}

\title{Below what training size do deep tabular generators stop beating trivial baselines?\\
\large A preregistered benchmark on a size ladder of clinical and standard datasets}
\author{Shivam Shrivastava\\
\small VIT Bhopal University, India\\
\small \texttt{shivamshrivastava2022@vitbhopal.ac.in}}
\date{\today}

\input{sections/numbers.tex}

\begin{document}
\maketitle

\begin{abstract}
\input{sections/abstract.tex}
\end{abstract}

\section{Introduction: the small-n regime and why clinical data lives there}
\input{sections/introduction.tex}

\section{Related work}
\input{sections/related.tex}

\section{Protocol}
\input{sections/protocol.tex}

\section{Results}
\input{sections/results.tex}

\section{Is subsampling a large dataset a valid proxy for a small one?}
\input{sections/subsample.tex}

\section{Limitations}
\input{sections/limitations.tex}

\section{Conclusion}
\input{sections/conclusion.tex}

\bibliographystyle{plainnat}
\bibliography{refs}

\appendix
\section{Per-cell results}
\input{tables/appendix_cells.tex}

\end{document}

%% file: sections/numbers.tex
\newcommand{\BaselineWinBeyondNoise}{28}
\newcommand{\BaselineWinCells}{40}
\newcommand{\BigRungWinFrac}{100}
\newcommand{\CeilingBeatCount}{2}
\newcommand{\CeilingGapMax}{0.017}

\newcommand{\CrossoverCount}{1}
\newcommand{\CrossoverDataset}{diabetes130}
\newcommand{\CrossoverMethod}{tvae}
\newcommand{\CrossoverN}{500}

\newcommand{\CtstCtgan}{0.947}

\newcommand{\CtstSmote}{0.940}
\newcommand{\CtstTabDDPM}{0.808}

\newcommand{\DatasetsAtFiveHundred}{8}
\newcommand{\DatasetsAtTwoHundred}{8}

\newcommand{\DcrCtgan}{0.529}

\newcommand{\DcrSmote}{0.925}
\newcommand{\DcrTabDDPM}{0.660}
\newcommand{\DcrTvae}{0.595}

\newcommand{\DeepWinsBeyondNoiseAtFiveHundred}{1}
\newcommand{\DeepWinsBeyondNoiseAtTwoHundred}{0}
\newcommand{\EffectCells}{53}
\newcommand{\EffectInsideNoise}{43}
\newcommand{\EffectInsideNoisePct}{81}
\newcommand{\FalsifyThresholdCount}{5}

\newcommand{\FitCtgan}{44.4}

\newcommand{\FitSmote}{0.1}
\newcommand{\FitTabDDPM}{104.3}

\newcommand{\MiaCtgan}{0.504}

\newcommand{\MiaSmote}{0.740}
\newcommand{\MiaTabDDPM}{0.567}
\newcommand{\MiaTvae}{0.530}

\newcommand{\NoCrossoverPairs}{23}
\newcommand{\NonOkCells}{0}
\newcommand{\NumCodeVersions}{1}

\newcommand{\NumLadderDatasets}{8}
\newcommand{\NumSeeds}{5}
\newcommand{\NumSmallDatasets}{4}
\newcommand{\NumTuningTrials}{20}
\newcommand{\OkCells}{2,220}
\newcommand{\SmallRungWinFrac}{62}
\newcommand{\StabilityThreshold}{0.8}
\newcommand{\SubsamplePairTauMax}{0.90}
\newcommand{\SubsamplePairTauMin}{0.24}
\newcommand{\SubsamplePairs}{32}
\newcommand{\SubsampleTauMax}{0.64}
\newcommand{\SubsampleTauMin}{0.57}

\newcommand{\TauDeepBelowFiveK}{0.806}
\newcommand{\TauDeepBelowFiveKStrict}{0.833}
\newcommand{\TauDeepLargestPair}{0.667}
\newcommand{\TauDeepLargestPairHi}{20000}
\newcommand{\TauDeepLargestPairLo}{10000}
\newcommand{\TauDeepSmallestPair}{0.917}
\newcommand{\TotalCells}{2,220}
\newcommand{\TotalCrossoverPairs}{24}
\newcommand{\TotalLadderCells}{49}

%% file: sections/abstract.tex
Deep tabular generative models are benchmarked on datasets with tens of
thousands of rows; clinical datasets have hundreds. We preregistered and
ran a size-ladder benchmark to find where the two regimes diverge:
\NumLadderDatasets{} public datasets subsampled from 200 to 20{,}000
training rows, seven generators (independent marginals, Gaussian copula,
SMOTE, unconditional SMOTE, CTGAN, TVAE, TabDDPM) with a fixed
\NumTuningTrials{}-trial tuning budget and \NumSeeds{} evaluation seeds,
plus \NumSmallDatasets{} natively small clinical datasets at true size,
for \TotalCells{} committed runs in total. The primary metric is the
AUROC of fixed classifiers trained on synthetic and tested on real data.

In \NoCrossoverPairs{} of \TotalCrossoverPairs{} (dataset, deep model)
pairs no deep model ever beats the best trivial baseline by more than
seed noise, at any training size we measured. The best baseline wins
\BaselineWinCells{} of \TotalLadderCells{} (dataset, size) cells. Our preregistered
prediction that the deep models' ranking would be unstable at small sizes
is falsified: mean Kendall $\tau$ between adjacent rungs below 5{,}000
rows is \TauDeepBelowFiveK{}, above our \StabilityThreshold{} threshold,
and stability is highest at the smallest sizes rather than lowest. One
caveat bounds all of this: in \EffectInsideNoisePct{}\% of cells the gap
between the top two methods is smaller than the variation between seeds. Finally, method rankings on
natively small clinical datasets agree only moderately with rankings on
subsampled large ones (mean $\tau$ \SubsampleTauMin{} to
\SubsampleTauMax{}), which questions whether a subsampled large dataset
can stand in for a small one.

All \OkCells{} result files, the preregistration and its hash, and the
code that regenerates every figure and number from those files are
public.

%% file: sections/introduction.tex
A clinical modelling team rarely has the data that tabular generative
models are benchmarked on. A single-site registry, a completed trial arm,
a rare-disease cohort: a few hundred to a few thousand rows, a dozen to a
hundred columns, a binary outcome, and a reason to share the data that the
data itself forbids. Synthetic data is the standard proposal for that
situation, and the models proposed for it are the same ones evaluated on
census tables with tens of thousands of rows: conditional GANs, tabular
VAEs, and more recently diffusion models. Benchmarks report that the
newest of these win, that tuning matters, and that the ranking between
them depends on the dataset. What they do not report, because their
datasets are large, is what happens at the sizes where clinical data
actually lives.

We ask two questions with one experiment. First, below what training-set
size do deep tabular generators stop beating trivial statistical
baselines, meaning independent marginals, a Gaussian copula, and SMOTE
interpolation, on the utility metric practitioners care about, the AUROC
of a classifier trained on synthetic data and tested on real data? Second,
in that regime, is the published ranking between the deep models stable,
or does it flip between adjacent training sizes and between random seeds?

The experiment is a size ladder. Eight public datasets are subsampled from
200 to 20{,}000 training rows, with validation and test sets held at full
size so that the only thing that shrinks is what the generator sees. Seven
generators run at every rung with a fixed tuning budget and five seeds, and
a disjoint sample of real rows sets the ceiling. Four natively small
clinical datasets are run at their true size, so that we can also ask
whether subsampling a large dataset is a valid proxy for a genuinely small
one.

The protocol was preregistered after a pilot and before any tuned run: the hypothesis, the primary metric, the search spaces, the analyses
and what would falsify the claim were committed and hashed. Every run,
including the ones that failed or timed out, is a committed result file,
and every number in this paper is regenerated from those files by one
command. We make no methodological contribution beyond the protocol; the
contribution is the measurement and the discipline around it.

%% file: sections/related.tex

Two recent studies frame ours. \citet{kindji2024tabular} benchmark five
model families on sixteen datasets averaging 80{,}000 rows with hundreds
of tuning trials per fold and report that diffusion models win when
tuning is unconstrained but lose their edge under a fixed GPU budget.
They also define the ucSMOTE baseline and the DCR-rate privacy measure we
adopt, and publish the search spaces we transcribe. Their question is
``which model, with enough data and tuning''; ours is ``below how many
rows does the question stop mattering'', asked on the same families with
the same baselines at sizes their datasets never reach.

\citet{du2024systematic} examine and critique the metrics used to
evaluate tabular synthesis and introduce fidelity, privacy and utility
measures across eight synthesizer types and twelve datasets. We take
from them the discipline of naming one primary metric and labelling the
rest secondary, and the joint reading of utility and privacy: a method
that wins utility by reproducing training rows has not won.

SMOTE \citep{chawla2002smote} predates every deep model here and is
routinely used as a strong tabular generator baseline; TabDDPM
\citep{kotelnikov2023tabddpm} showed a simple interpolation baseline
competitive with several GANs. Our contribution is not a new model or
metric but a preregistered protocol, a size ladder that reaches the
regime of a single clinical study, a natively small arm to test whether
subsampling is a valid proxy for it, and committed result files for every
run, including the failures.

%% file: sections/protocol.tex
\label{sec:protocol}

\paragraph{Preregistration.}
The hypothesis, primary metric, datasets, size ladder, methods, seed
count, tuning budget, search spaces, analyses and falsification rule were
committed to the repository after a pilot and before any tuned run.
The frozen text and its hash are in the repository; every later change is
an appended, dated entry.

\paragraph{Hypothesis.}
Below 5{,}000 training rows, no deep tabular generative model (CTGAN,
TVAE, TabDDPM), even after a 20-trial tuning budget, achieves a higher
TSTR AUROC than the best of four trivial baselines (independent
marginals, Gaussian copula, SMOTE, ucSMOTE), and the ranking among the
deep models at those sizes is not stable between adjacent rungs. The
threshold was set by the pilot.

\paragraph{Datasets and splits.}
Eight ladder datasets are subsampled down a ladder of 200, 500, 1000,
2000, 5000, 10{,}000 and 20{,}000 training rows, capped at each
dataset's training split; four natively small clinical
datasets are used at their true size and never subsampled. Splits are
stratified 60/20/20, seeded by dataset id; validation and test stay at
full size at every rung; numeric missing values are imputed with the
training split's medians and categorical ones become an explicit level.
Sources, checksums, target definitions and cleaning are in the dataset
registry.

\paragraph{Methods.}
Baselines: independent empirical marginals (the floor), Gaussian copula,
SMOTE conditioned on the target, and unconditional ucSMOTE. Deep models:
CTGAN, TVAE, TabDDPM. A disjoint sample of real training rows is the
reference ceiling and never competes in any ranking. All implement one
interface and the runner treats them identically.

\paragraph{Primary metric.}
TSTR AUROC: logistic regression, random forest and XGBoost with fixed
constants are trained on the synthetic frame and scored on the full real
test split; the three AUROCs are averaged. Everything else (accuracy, F1,
TRTR, C2ST, KS, TVD, association distance, DCR, NNDR, membership
inference, cost) is secondary.

\paragraph{Tuning.}
Per (dataset, $n$, method): 20 Optuna TPE trials (19 grid points for the
SMOTE neighbour count) with a fixed tuning seed, objective the primary
metric on the validation split; the test split is never read during
tuning and the code asserts it. The winning configuration is then run
with five evaluation seeds. Search spaces are the reduced spaces of
\citet{kindji2024tabular} with declared compute-bound adaptations for
TabDDPM (at most four layers of 512 units, batch at most 1{,}024, 10{,}000
training iterations) and library defaults where SDV does not expose a
parameter.

\paragraph{Analyses.}
(1) Crossover: per dataset and deep method, the smallest $n$ at which the
deep model's mean exceeds the best baseline's mean by more than their
pooled seed standard deviation. (2) Headline: per rung, the fraction of
datasets in which the best baseline beats every deep model. (3) Ranking
stability: Kendall $\tau$ between rankings at adjacent rungs. (4) Effect
size against noise: the top-two gap divided by the pair standard
deviation. (5) Subsample validity: Kendall $\tau$ between each small
dataset's ranking and the ladder rankings at the nearest rung.

\paragraph{Reproducibility.}
Every result is one JSON file with the git SHA and a hash of the
run-producing code, versions, seeds, hyper-parameters, status and
traceback on failure. Failed and timed-out runs are recorded, never
dropped. Every table and figure in this paper is regenerated by one
command from those files.

%% file: sections/results.tex
\label{sec:results}

Two units recur below. A \emph{run} is one (dataset, $n$, method, seed)
combination and produces one committed result file. A \emph{cell} is one
(dataset, $n$) pair, the unit at which the preregistered analyses compare
methods.

All \TotalCells{} planned runs completed: \OkCells{} with status
\texttt{ok} and \NonOkCells{} failures, timeouts or out-of-memory
terminations, across \NumCodeVersions{} version of the run-producing
code. No run was dropped, and none needed to be.

\subsection{Utility against training size}

\begin{figure}[t]
  \centering
  \includegraphics[width=\linewidth]{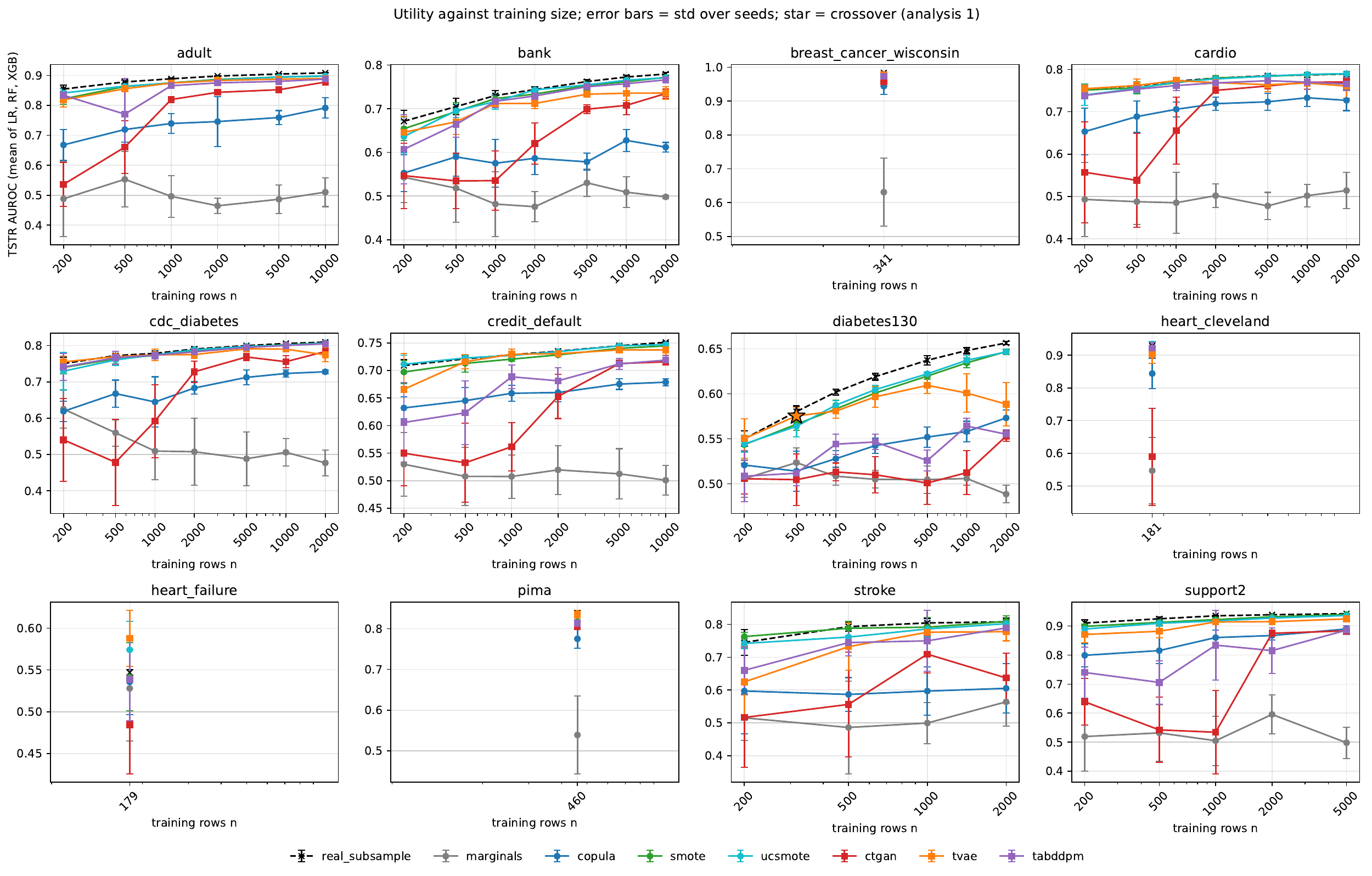}
  \caption{TSTR AUROC (mean of logistic regression, random forest and
  XGBoost) against training rows, one line per method, error bars one
  standard deviation over \NumSeeds{} seeds. Dashed black: a disjoint
  sample of real rows, the ceiling. Stars would mark crossover points
  (analysis~1); there is \CrossoverCount{}.}
  \label{fig:ladder}
\end{figure}

\input{tables/main.tex}

Figure~\ref{fig:ladder} and Table~\ref{tab:main} carry the main result.
In \NoCrossoverPairs{} of \TotalCrossoverPairs{} (ladder dataset, deep
model) pairs, the deep model never exceeds the best trivial baseline by
more than their pooled seed standard deviation at any rung of the ladder.
The single exception is \CrossoverMethod{} on \CrossoverDataset{} at
$n = \CrossoverN{}$, and it does not persist: the baselines are ahead
again at every larger rung of that dataset.

The baselines also sit close to the ceiling. At $n = 200$ the best
baseline is never more than \CeilingGapMax{} AUROC below real training
rows on any ladder dataset, and on \CeilingBeatCount{} of them it is
above real rows outright. That is not
an artefact: with a few hundred rows and a rare positive class, SMOTE's
interpolation between real minority cases gives a downstream classifier
more usable signal than the handful of raw rows it interpolated from.

\subsection{Headline: how often a trivial baseline wins}

\begin{figure}[t]
  \centering
  \includegraphics[width=0.6\linewidth]{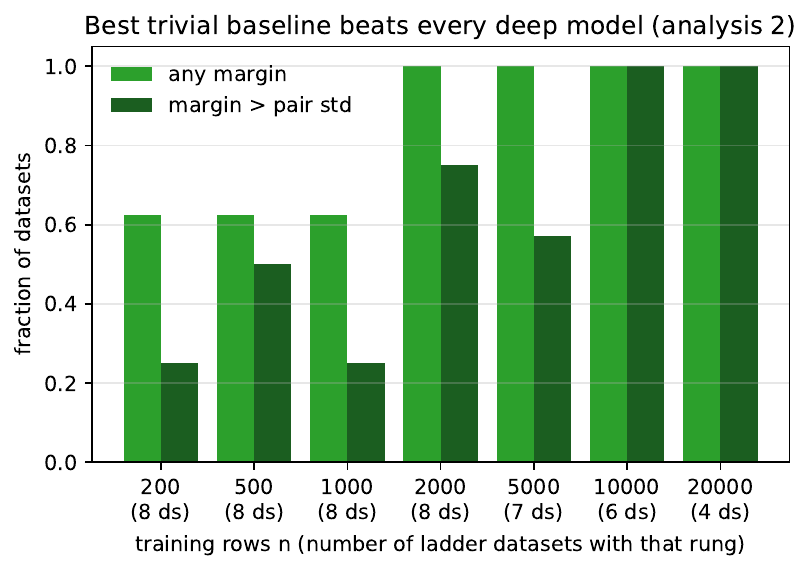}
  \caption{Fraction of ladder datasets, per rung, in which the best
  trivial baseline beats every deep model on the primary metric: any
  margin (light) and margin larger than the pair standard deviation
  (dark). Analysis~2.}
  \label{fig:headline}
\end{figure}

Over the whole ladder the best baseline beats every deep model in
\BaselineWinCells{} of \TotalLadderCells{} (dataset, $n$) cells, and in
\BaselineWinBeyondNoise{} of them the margin exceeds the pair standard
deviation.

Split by rung, the direction is the opposite of what we assumed when we
preregistered. The baseline wins in \SmallRungWinFrac{}\% of datasets at
the three smallest rungs and in \BigRungWinFrac{}\% at every rung from
2{,}000 rows upward (Figure~\ref{fig:headline}). We expected the baseline
advantage to be largest where data is scarcest. Instead, scarce data is
where the best deep model and the best baseline are hardest to tell
apart, and the baseline's advantage becomes unambiguous only once there
is more data, at exactly the sizes where practitioners would reach for a
deep model with most confidence.

\subsection{Preregistered falsification}

The preregistration states two falsification conditions, and the results
split between them. We report both as found.

\paragraph{The main hypothesis is not falsified.}
The condition was that a tuned deep model beats the best tuned baseline
beyond the pair standard deviation in \FalsifyThresholdCount{} or more of
the \NumLadderDatasets{} ladder datasets at $n \le 500$. It does so in
\DeepWinsBeyondNoiseAtTwoHundred{} of \DatasetsAtTwoHundred{} datasets at $n = 200$ and
\DeepWinsBeyondNoiseAtFiveHundred{} of \DatasetsAtFiveHundred{} at $n = 500$.

\paragraph{The ranking-stability clause is falsified.}
We predicted that the ranking among the three deep models would be
unstable between adjacent rungs at small sizes, and preregistered
falsification at a mean Kendall $\tau$ of \StabilityThreshold{} or above.
The measured mean is \TauDeepBelowFiveK{} counting every rung pair up to
5{,}000 rows, and \TauDeepBelowFiveKStrict{} on the stricter reading that
both rungs lie below 5{,}000. The clause is falsified on either reading,
so the ambiguity in our own wording does not need resolving.

\begin{figure}[t]
  \centering
  \includegraphics[width=\linewidth]{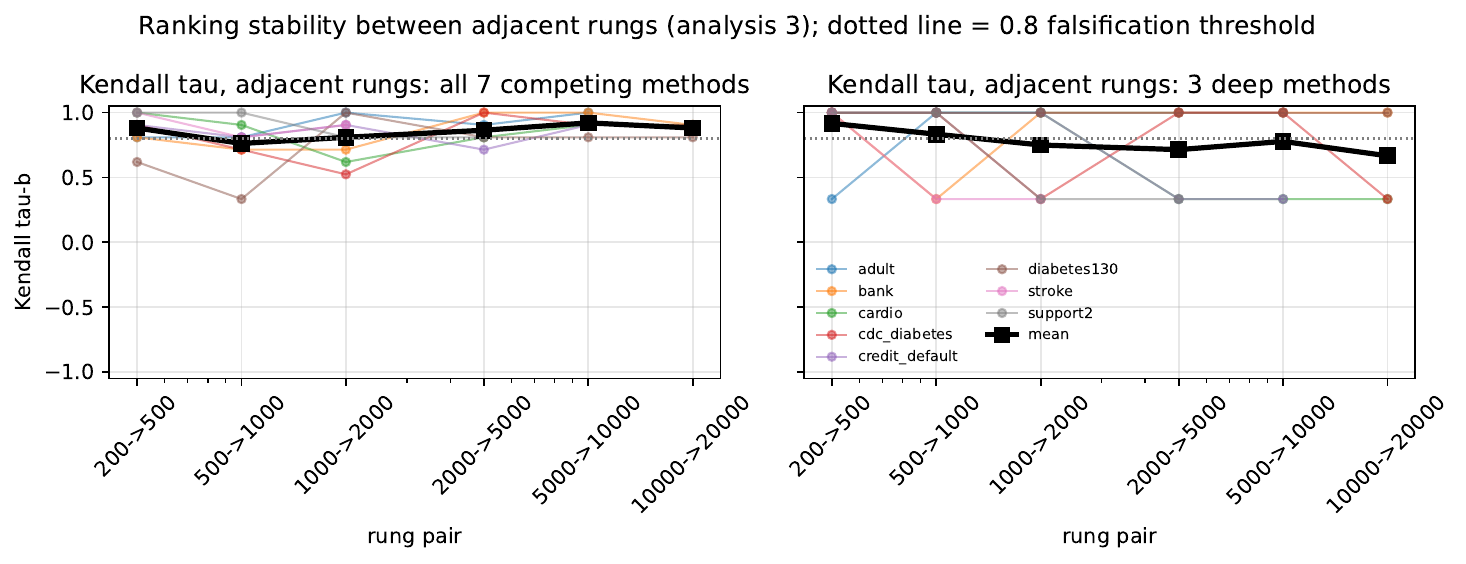}
  \caption{Kendall $\tau$ between method rankings at adjacent rungs, per
  dataset (thin) and mean (thick), for all seven competing methods (left)
  and the three deep models (right). Dotted line: the
  \StabilityThreshold{} threshold of the preregistered falsification
  rule. Analysis~3.}
  \label{fig:stability}
\end{figure}

Stability is \emph{highest} at the smallest rungs, $\tau = \TauDeepSmallestPair{}$
between 200 and 500 rows, and decays as data grows, reaching
$\tau = \TauDeepLargestPair{}$ between \TauDeepLargestPairLo{} and
\TauDeepLargestPairHi{} rows (Figure~\ref{fig:stability}). The deep
models keep a consistent order at small $n$ because all three are
performing poorly in the same way; they diverge later, once there is
enough data for their architectural differences to express themselves.

\subsection{Effect size against seed noise}

\begin{figure}[t]
  \centering
  \includegraphics[width=0.55\linewidth]{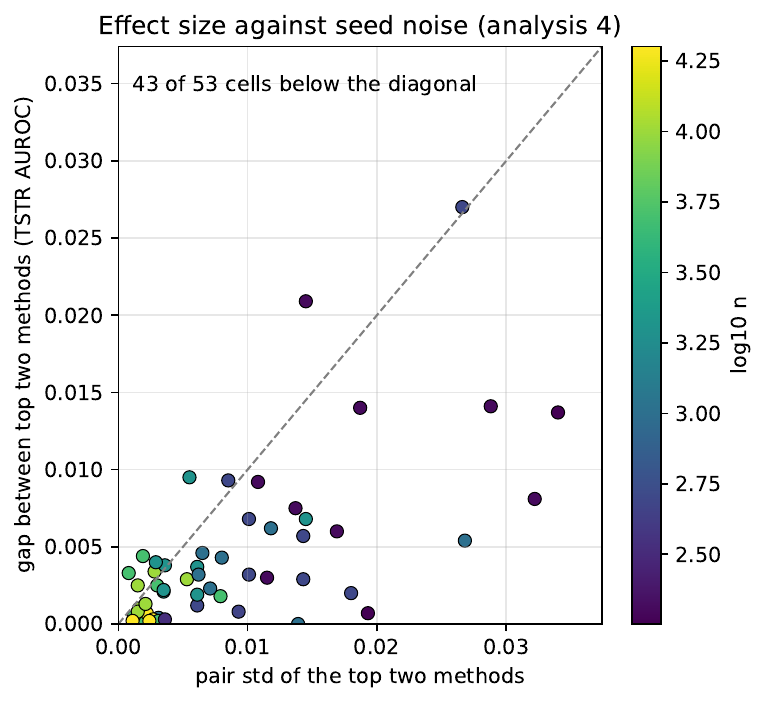}
  \caption{Gap between the top two competing methods against their pair
  standard deviation, one point per (dataset, $n$). Below the diagonal
  the apparent winner is inside the noise. Analysis~4.}
  \label{fig:effect}
\end{figure}

In \EffectInsideNoise{} of \EffectCells{} cells, that is
\EffectInsideNoisePct{}\%, the gap between the best and second-best
method is smaller than the pooled standard deviation of those two methods
over \NumSeeds{} seeds (Figure~\ref{fig:effect}). Part of that is the two
SMOTE variants trading places with each other, which is of no
consequence. But it also means that a benchmark table reporting a single
seed, as many do, would frequently name a different winner on a rerun.

The defensible claim from this study is therefore a narrow one. At small
$n$ there is no measurable reason to prefer a deep generator, at a cost
difference of \FitTabDDPM{} seconds per fit for TabDDPM and \FitCtgan{}
for CTGAN against \FitSmote{} seconds for SMOTE. The wider claim, that
trivial baselines are better, is more than these margins will carry.

\subsection{Utility and privacy jointly}

\begin{figure}[t]
  \centering
  \includegraphics[width=\linewidth]{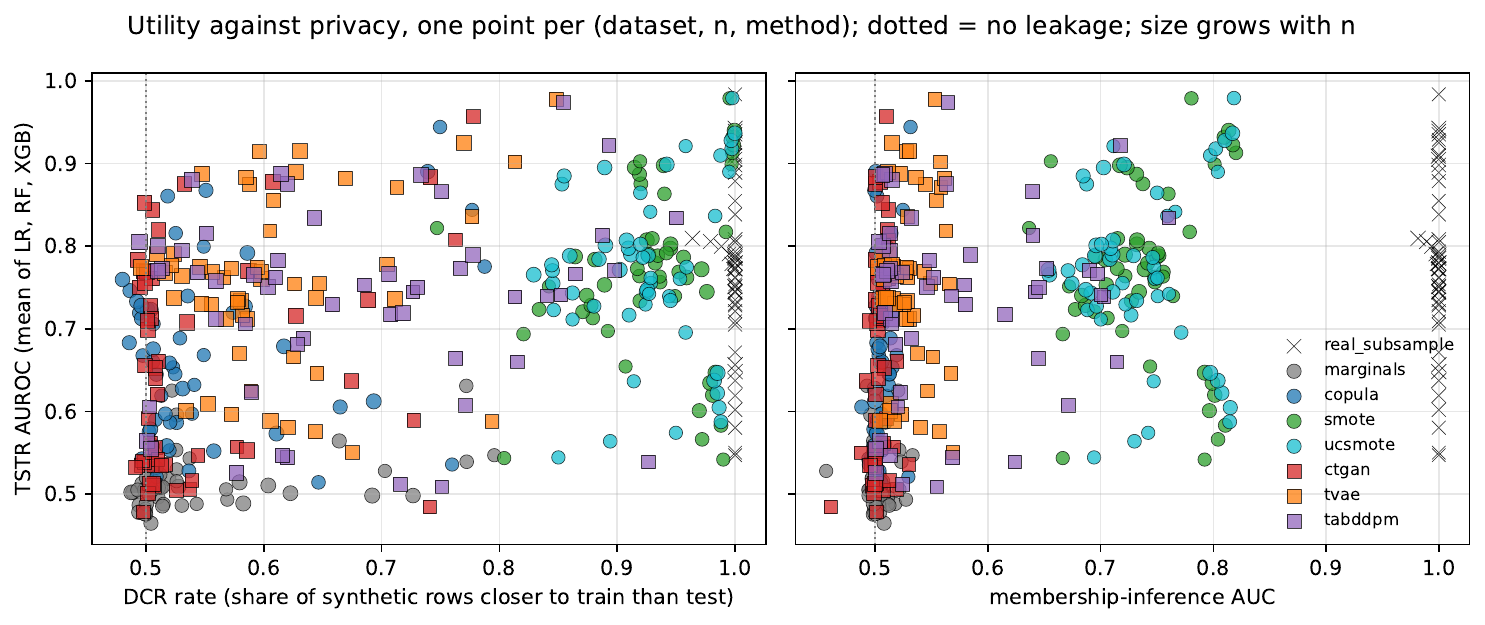}
  \caption{Primary metric against DCR rate (left) and
  membership-inference AUC (right), one point per (dataset, $n$, method);
  marker size grows with $n$. Dotted lines mark the no-leakage value 0.5;
  real rows sit near 1.0 on both axes. Secondary and descriptive.}
  \label{fig:privacy}
\end{figure}

A method that wins on utility by copying its training data has not won,
so the two axes have to be read together.

Averaged over the ladder cells, SMOTE's DCR rate is \DcrSmote{} and its
membership-inference AUC is \MiaSmote{}, against \DcrTvae{} and
\MiaTvae{} for TVAE, \DcrTabDDPM{} and \MiaTabDDPM{} for TabDDPM, and
\DcrCtgan{} and \MiaCtgan{} for CTGAN; 0.5 is the no-leakage value on
both. The methods that win utility are the ones that stay closest to
real records, which is the same property that makes them leak. Anyone
adopting a SMOTE variant on the strength of Section~\ref{sec:results}
inherits that trade-off, and it is the strongest argument in this study
for continuing to develop the deep models rather than abandoning them.

The fidelity metric points the same way from the other side. TabDDPM
produces the least distinguishable synthetic data of any generator, with
a mean C2ST AUC of \CtstTabDDPM{} where 0.5 is indistinguishable, better
than SMOTE's \CtstSmote{} and CTGAN's \CtstCtgan{}. It is the most
realistic generator in the study and still not the most useful one.
Fidelity and utility are not the same axis, and a benchmark that reports
only one of them will mislead.

%% file: tables/main.tex
\begin{table}[t]
\centering
\small
\caption{TSTR AUROC (mean $\pm$ std over 5 seeds) per dataset and training size. Bold: best competing method (real\_subsample is the reference ceiling).}
\label{tab:main}
\resizebox{\textwidth}{!}{%
\begin{tabular}{llllllllll}
\toprule
dataset & n & real\_subsample & marginals & copula & smote & ucsmote & ctgan & tvae & tabddpm \\
\midrule
adult & 200 & 0.855 $\pm$ 0.013 & 0.488 $\pm$ 0.126 & 0.668 $\pm$ 0.052 & 0.822 $\pm$ 0.019 & \textbf{0.841 $\pm$ 0.015} & 0.536 $\pm$ 0.073 & 0.819 $\pm$ 0.023 & 0.834 $\pm$ 0.013 \\
adult & 500 & 0.879 $\pm$ 0.006 & 0.553 $\pm$ 0.093 & 0.720 $\pm$ 0.042 & 0.863 $\pm$ 0.005 & \textbf{0.864 $\pm$ 0.007} & 0.661 $\pm$ 0.088 & 0.856 $\pm$ 0.010 & 0.771 $\pm$ 0.119 \\
adult & 1000 & 0.889 $\pm$ 0.002 & 0.496 $\pm$ 0.070 & 0.740 $\pm$ 0.033 & \textbf{0.875 $\pm$ 0.004} & 0.875 $\pm$ 0.002 & 0.820 $\pm$ 0.006 & 0.875 $\pm$ 0.008 & 0.866 $\pm$ 0.006 \\
adult & 2000 & 0.898 $\pm$ 0.001 & 0.465 $\pm$ 0.026 & 0.746 $\pm$ 0.084 & \textbf{0.887 $\pm$ 0.003} & 0.885 $\pm$ 0.004 & 0.844 $\pm$ 0.008 & 0.883 $\pm$ 0.003 & 0.875 $\pm$ 0.005 \\
adult & 5000 & 0.905 $\pm$ 0.001 & 0.486 $\pm$ 0.048 & 0.760 $\pm$ 0.023 & 0.895 $\pm$ 0.002 & \textbf{0.895 $\pm$ 0.001} & 0.852 $\pm$ 0.003 & 0.888 $\pm$ 0.007 & 0.880 $\pm$ 0.004 \\
adult & 10000 & 0.909 $\pm$ 0.001 & 0.510 $\pm$ 0.048 & 0.792 $\pm$ 0.033 & 0.898 $\pm$ 0.002 & \textbf{0.899 $\pm$ 0.001} & 0.878 $\pm$ 0.004 & 0.890 $\pm$ 0.002 & 0.888 $\pm$ 0.002 \\
bank & 200 & 0.671 $\pm$ 0.025 & 0.543 $\pm$ 0.057 & 0.553 $\pm$ 0.042 & \textbf{0.654 $\pm$ 0.027} & 0.636 $\pm$ 0.038 & 0.547 $\pm$ 0.075 & 0.646 $\pm$ 0.037 & 0.607 $\pm$ 0.079 \\
bank & 500 & 0.705 $\pm$ 0.019 & 0.518 $\pm$ 0.078 & 0.590 $\pm$ 0.045 & 0.693 $\pm$ 0.021 & \textbf{0.695 $\pm$ 0.015} & 0.535 $\pm$ 0.063 & 0.670 $\pm$ 0.029 & 0.664 $\pm$ 0.029 \\
bank & 1000 & 0.730 $\pm$ 0.011 & 0.482 $\pm$ 0.074 & 0.575 $\pm$ 0.055 & \textbf{0.723 $\pm$ 0.008} & 0.716 $\pm$ 0.018 & 0.535 $\pm$ 0.068 & 0.712 $\pm$ 0.009 & 0.717 $\pm$ 0.015 \\
bank & 2000 & 0.744 $\pm$ 0.006 & 0.476 $\pm$ 0.035 & 0.587 $\pm$ 0.037 & 0.734 $\pm$ 0.006 & \textbf{0.743 $\pm$ 0.005} & 0.620 $\pm$ 0.047 & 0.712 $\pm$ 0.011 & 0.729 $\pm$ 0.009 \\
bank & 5000 & 0.762 $\pm$ 0.004 & 0.530 $\pm$ 0.031 & 0.578 $\pm$ 0.020 & 0.753 $\pm$ 0.010 & \textbf{0.755 $\pm$ 0.005} & 0.699 $\pm$ 0.010 & 0.733 $\pm$ 0.006 & 0.750 $\pm$ 0.005 \\
bank & 10000 & 0.773 $\pm$ 0.004 & 0.509 $\pm$ 0.035 & 0.628 $\pm$ 0.025 & 0.762 $\pm$ 0.003 & \textbf{0.765 $\pm$ 0.003} & 0.708 $\pm$ 0.021 & 0.736 $\pm$ 0.017 & 0.757 $\pm$ 0.006 \\
bank & 20000 & 0.780 $\pm$ 0.003 & 0.498 $\pm$ 0.003 & 0.612 $\pm$ 0.011 & \textbf{0.772 $\pm$ 0.003} & 0.771 $\pm$ 0.001 & 0.735 $\pm$ 0.011 & 0.736 $\pm$ 0.014 & 0.766 $\pm$ 0.006 \\
breast\_cancer\_wisconsin & 341 & 0.984 $\pm$ 0.000 & 0.631 $\pm$ 0.101 & 0.944 $\pm$ 0.024 & 0.979 $\pm$ 0.003 & \textbf{0.979 $\pm$ 0.004} & 0.957 $\pm$ 0.013 & 0.978 $\pm$ 0.005 & 0.974 $\pm$ 0.006 \\
cardio & 200 & 0.754 $\pm$ 0.008 & 0.493 $\pm$ 0.087 & 0.653 $\pm$ 0.055 & 0.751 $\pm$ 0.014 & 0.739 $\pm$ 0.025 & 0.557 $\pm$ 0.119 & \textbf{0.754 $\pm$ 0.008} & 0.739 $\pm$ 0.008 \\
cardio & 500 & 0.761 $\pm$ 0.009 & 0.488 $\pm$ 0.054 & 0.689 $\pm$ 0.037 & 0.756 $\pm$ 0.014 & 0.755 $\pm$ 0.010 & 0.538 $\pm$ 0.111 & \textbf{0.762 $\pm$ 0.015} & 0.753 $\pm$ 0.005 \\
cardio & 1000 & 0.772 $\pm$ 0.003 & 0.485 $\pm$ 0.072 & 0.706 $\pm$ 0.018 & 0.769 $\pm$ 0.003 & 0.770 $\pm$ 0.004 & 0.656 $\pm$ 0.079 & \textbf{0.773 $\pm$ 0.008} & 0.762 $\pm$ 0.012 \\
cardio & 2000 & 0.779 $\pm$ 0.002 & 0.502 $\pm$ 0.028 & 0.719 $\pm$ 0.016 & \textbf{0.779 $\pm$ 0.001} & 0.777 $\pm$ 0.005 & 0.750 $\pm$ 0.003 & 0.769 $\pm$ 0.008 & 0.768 $\pm$ 0.012 \\
cardio & 5000 & 0.785 $\pm$ 0.001 & 0.478 $\pm$ 0.032 & 0.723 $\pm$ 0.020 & 0.784 $\pm$ 0.001 & \textbf{0.784 $\pm$ 0.002} & 0.761 $\pm$ 0.006 & 0.764 $\pm$ 0.016 & 0.773 $\pm$ 0.016 \\
cardio & 10000 & 0.787 $\pm$ 0.005 & 0.502 $\pm$ 0.027 & 0.733 $\pm$ 0.020 & 0.787 $\pm$ 0.001 & \textbf{0.788 $\pm$ 0.002} & 0.770 $\pm$ 0.005 & 0.768 $\pm$ 0.014 & 0.770 $\pm$ 0.017 \\
cardio & 20000 & 0.789 $\pm$ 0.002 & 0.514 $\pm$ 0.043 & 0.727 $\pm$ 0.024 & 0.790 $\pm$ 0.001 & \textbf{0.790 $\pm$ 0.001} & 0.770 $\pm$ 0.008 & 0.761 $\pm$ 0.028 & 0.765 $\pm$ 0.016 \\
cdc\_diabetes & 200 & 0.750 $\pm$ 0.018 & 0.625 $\pm$ 0.053 & 0.619 $\pm$ 0.028 & 0.739 $\pm$ 0.017 & 0.729 $\pm$ 0.052 & 0.540 $\pm$ 0.115 & \textbf{0.755 $\pm$ 0.016} & 0.741 $\pm$ 0.037 \\
cdc\_diabetes & 500 & 0.772 $\pm$ 0.006 & 0.560 $\pm$ 0.036 & 0.668 $\pm$ 0.037 & 0.764 $\pm$ 0.014 & 0.760 $\pm$ 0.015 & 0.478 $\pm$ 0.118 & \textbf{0.769 $\pm$ 0.009} & 0.766 $\pm$ 0.018 \\
cdc\_diabetes & 1000 & 0.778 $\pm$ 0.007 & 0.509 $\pm$ 0.078 & 0.645 $\pm$ 0.069 & 0.773 $\pm$ 0.007 & 0.774 $\pm$ 0.011 & 0.592 $\pm$ 0.101 & \textbf{0.774 $\pm$ 0.016} & 0.773 $\pm$ 0.010 \\
cdc\_diabetes & 2000 & 0.790 $\pm$ 0.005 & 0.508 $\pm$ 0.092 & 0.683 $\pm$ 0.017 & 0.787 $\pm$ 0.007 & \textbf{0.788 $\pm$ 0.005} & 0.728 $\pm$ 0.030 & 0.775 $\pm$ 0.011 & 0.783 $\pm$ 0.007 \\
cdc\_diabetes & 5000 & 0.800 $\pm$ 0.002 & 0.488 $\pm$ 0.074 & 0.712 $\pm$ 0.020 & 0.797 $\pm$ 0.003 & \textbf{0.797 $\pm$ 0.003} & 0.768 $\pm$ 0.010 & 0.791 $\pm$ 0.004 & 0.795 $\pm$ 0.004 \\
cdc\_diabetes & 10000 & 0.806 $\pm$ 0.001 & 0.506 $\pm$ 0.038 & 0.723 $\pm$ 0.010 & \textbf{0.802 $\pm$ 0.001} & 0.801 $\pm$ 0.003 & 0.755 $\pm$ 0.016 & 0.790 $\pm$ 0.006 & 0.800 $\pm$ 0.001 \\
cdc\_diabetes & 20000 & 0.809 $\pm$ 0.002 & 0.477 $\pm$ 0.036 & 0.728 $\pm$ 0.005 & \textbf{0.807 $\pm$ 0.002} & 0.807 $\pm$ 0.003 & 0.783 $\pm$ 0.006 & 0.774 $\pm$ 0.019 & 0.805 $\pm$ 0.002 \\
credit\_default & 200 & 0.709 $\pm$ 0.011 & 0.530 $\pm$ 0.058 & 0.632 $\pm$ 0.020 & 0.697 $\pm$ 0.021 & \textbf{0.711 $\pm$ 0.016} & 0.550 $\pm$ 0.059 & 0.666 $\pm$ 0.065 & 0.606 $\pm$ 0.072 \\
credit\_default & 500 & 0.721 $\pm$ 0.006 & 0.508 $\pm$ 0.052 & 0.645 $\pm$ 0.024 & 0.713 $\pm$ 0.016 & \textbf{0.723 $\pm$ 0.005} & 0.533 $\pm$ 0.072 & 0.716 $\pm$ 0.013 & 0.623 $\pm$ 0.058 \\
credit\_default & 1000 & 0.728 $\pm$ 0.003 & 0.508 $\pm$ 0.039 & 0.659 $\pm$ 0.015 & 0.721 $\pm$ 0.004 & 0.727 $\pm$ 0.004 & 0.562 $\pm$ 0.044 & \textbf{0.729 $\pm$ 0.009} & 0.688 $\pm$ 0.021 \\
credit\_default & 2000 & 0.735 $\pm$ 0.004 & 0.520 $\pm$ 0.044 & 0.660 $\pm$ 0.017 & 0.728 $\pm$ 0.002 & \textbf{0.734 $\pm$ 0.004} & 0.653 $\pm$ 0.040 & 0.730 $\pm$ 0.003 & 0.681 $\pm$ 0.024 \\
credit\_default & 5000 & 0.745 $\pm$ 0.002 & 0.513 $\pm$ 0.046 & 0.675 $\pm$ 0.010 & 0.740 $\pm$ 0.002 & \textbf{0.745 $\pm$ 0.002} & 0.713 $\pm$ 0.006 & 0.737 $\pm$ 0.006 & 0.712 $\pm$ 0.010 \\
credit\_default & 10000 & 0.751 $\pm$ 0.002 & 0.501 $\pm$ 0.027 & 0.679 $\pm$ 0.006 & 0.745 $\pm$ 0.001 & \textbf{0.747 $\pm$ 0.002} & 0.716 $\pm$ 0.002 & 0.737 $\pm$ 0.010 & 0.719 $\pm$ 0.008 \\
diabetes130 & 200 & 0.550 $\pm$ 0.009 & 0.506 $\pm$ 0.021 & 0.521 $\pm$ 0.016 & 0.543 $\pm$ 0.007 & 0.544 $\pm$ 0.009 & 0.506 $\pm$ 0.017 & \textbf{0.550 $\pm$ 0.022} & 0.508 $\pm$ 0.028 \\
diabetes130 & 500 & 0.580 $\pm$ 0.006 & 0.524 $\pm$ 0.016 & 0.514 $\pm$ 0.022 & 0.566 $\pm$ 0.007 & 0.564 $\pm$ 0.012 & 0.505 $\pm$ 0.028 & \textbf{0.575 $\pm$ 0.010} & 0.512 $\pm$ 0.014 \\
diabetes130 & 1000 & 0.602 $\pm$ 0.003 & 0.508 $\pm$ 0.010 & 0.528 $\pm$ 0.009 & 0.583 $\pm$ 0.010 & \textbf{0.587 $\pm$ 0.005} & 0.513 $\pm$ 0.010 & 0.581 $\pm$ 0.008 & 0.544 $\pm$ 0.011 \\
diabetes130 & 2000 & 0.619 $\pm$ 0.004 & 0.505 $\pm$ 0.010 & 0.542 $\pm$ 0.008 & 0.601 $\pm$ 0.007 & \textbf{0.605 $\pm$ 0.005} & 0.510 $\pm$ 0.020 & 0.597 $\pm$ 0.012 & 0.546 $\pm$ 0.008 \\
diabetes130 & 5000 & 0.637 $\pm$ 0.005 & 0.505 $\pm$ 0.015 & 0.552 $\pm$ 0.011 & 0.620 $\pm$ 0.004 & \textbf{0.622 $\pm$ 0.002} & 0.501 $\pm$ 0.024 & 0.609 $\pm$ 0.009 & 0.526 $\pm$ 0.012 \\
diabetes130 & 10000 & 0.648 $\pm$ 0.004 & 0.506 $\pm$ 0.008 & 0.558 $\pm$ 0.011 & 0.634 $\pm$ 0.006 & \textbf{0.637 $\pm$ 0.005} & 0.512 $\pm$ 0.024 & 0.601 $\pm$ 0.021 & 0.564 $\pm$ 0.009 \\
diabetes130 & 20000 & 0.656 $\pm$ 0.002 & 0.489 $\pm$ 0.010 & 0.573 $\pm$ 0.009 & \textbf{0.647 $\pm$ 0.003} & 0.647 $\pm$ 0.002 & 0.553 $\pm$ 0.006 & 0.589 $\pm$ 0.024 & 0.555 $\pm$ 0.005 \\
heart\_cleveland & 181 & 0.929 $\pm$ 0.001 & 0.547 $\pm$ 0.101 & 0.844 $\pm$ 0.046 & 0.903 $\pm$ 0.006 & 0.921 $\pm$ 0.021 & 0.589 $\pm$ 0.148 & 0.902 $\pm$ 0.027 & \textbf{0.922 $\pm$ 0.017} \\
heart\_failure & 179 & 0.547 $\pm$ 0.004 & 0.528 $\pm$ 0.063 & 0.536 $\pm$ 0.039 & 0.542 $\pm$ 0.041 & 0.574 $\pm$ 0.034 & 0.485 $\pm$ 0.059 & \textbf{0.588 $\pm$ 0.034} & 0.539 $\pm$ 0.050 \\
pima & 460 & 0.838 $\pm$ 0.002 & 0.539 $\pm$ 0.096 & 0.775 $\pm$ 0.023 & 0.817 $\pm$ 0.011 & \textbf{0.836 $\pm$ 0.009} & 0.808 $\pm$ 0.010 & 0.836 $\pm$ 0.010 & 0.813 $\pm$ 0.009 \\
stroke & 200 & 0.745 $\pm$ 0.039 & 0.516 $\pm$ 0.069 & 0.597 $\pm$ 0.130 & \textbf{0.763 $\pm$ 0.013} & 0.742 $\pm$ 0.015 & 0.517 $\pm$ 0.153 & 0.625 $\pm$ 0.110 & 0.660 $\pm$ 0.071 \\
stroke & 500 & 0.793 $\pm$ 0.014 & 0.486 $\pm$ 0.142 & 0.587 $\pm$ 0.051 & \textbf{0.788 $\pm$ 0.022} & 0.761 $\pm$ 0.031 & 0.556 $\pm$ 0.159 & 0.732 $\pm$ 0.072 & 0.745 $\pm$ 0.040 \\
stroke & 1000 & 0.805 $\pm$ 0.014 & 0.500 $\pm$ 0.063 & 0.597 $\pm$ 0.074 & \textbf{0.792 $\pm$ 0.030} & 0.786 $\pm$ 0.023 & 0.709 $\pm$ 0.057 & 0.776 $\pm$ 0.018 & 0.750 $\pm$ 0.093 \\
stroke & 2000 & 0.808 $\pm$ 0.012 & 0.564 $\pm$ 0.074 & 0.605 $\pm$ 0.075 & \textbf{0.809 $\pm$ 0.018} & 0.802 $\pm$ 0.010 & 0.637 $\pm$ 0.076 & 0.778 $\pm$ 0.028 & 0.789 $\pm$ 0.013 \\
support2 & 200 & 0.910 $\pm$ 0.011 & 0.519 $\pm$ 0.119 & 0.799 $\pm$ 0.039 & \textbf{0.899 $\pm$ 0.008} & 0.890 $\pm$ 0.013 & 0.639 $\pm$ 0.081 & 0.871 $\pm$ 0.029 & 0.740 $\pm$ 0.088 \\
support2 & 500 & 0.925 $\pm$ 0.007 & 0.532 $\pm$ 0.098 & 0.816 $\pm$ 0.044 & \textbf{0.913 $\pm$ 0.009} & 0.910 $\pm$ 0.011 & 0.542 $\pm$ 0.113 & 0.882 $\pm$ 0.022 & 0.706 $\pm$ 0.075 \\
support2 & 1000 & 0.935 $\pm$ 0.002 & 0.505 $\pm$ 0.085 & 0.860 $\pm$ 0.026 & \textbf{0.922 $\pm$ 0.008} & 0.918 $\pm$ 0.005 & 0.534 $\pm$ 0.144 & 0.914 $\pm$ 0.006 & 0.834 $\pm$ 0.120 \\
support2 & 2000 & 0.939 $\pm$ 0.001 & 0.596 $\pm$ 0.067 & 0.867 $\pm$ 0.016 & \textbf{0.932 $\pm$ 0.003} & 0.928 $\pm$ 0.002 & 0.875 $\pm$ 0.004 & 0.915 $\pm$ 0.005 & 0.815 $\pm$ 0.078 \\
support2 & 5000 & 0.943 $\pm$ 0.001 & 0.498 $\pm$ 0.054 & 0.890 $\pm$ 0.003 & \textbf{0.940 $\pm$ 0.001} & 0.936 $\pm$ 0.001 & 0.883 $\pm$ 0.003 & 0.925 $\pm$ 0.002 & 0.886 $\pm$ 0.016 \\
\bottomrule
\end{tabular}
}
\end{table}

%% file: sections/subsample.tex
\label{sec:subsample}

Small-data comparisons are commonly built by subsampling a large public
dataset, the ladder in Section~\ref{sec:results} included. That shortcut
assumes a 200-row draw from a 250{,}000-row survey
behaves like a 200-row clinical study. Stage~2 tests the assumption
directly.

We ran the same \NumSmallDatasets{} natively small clinical datasets at
their true training sizes, with the same methods, the same
\NumTuningTrials{}-trial tuning and the same \NumSeeds{} seeds, and
compared each one's method ranking against the ranking of every ladder
dataset at the nearest rung, by Kendall $\tau$ (analysis~5).

Agreement is moderate at best. Mean $\tau$ per small dataset ranges from
\SubsampleTauMin{} to \SubsampleTauMax{}, and across the
\SubsamplePairs{} individual pairs it ranges from \SubsamplePairTauMin{}
to \SubsamplePairTauMax{}. A $\tau$ near \SubsampleTauMin{} over seven
competing methods means several adjacent swaps: the top method usually
survives, the middle of the ranking does not.

The per-dataset results show why the averages hide trouble. On the ladder
the ordering is uniform, with a SMOTE variant first almost everywhere. On
the natively small sets it is not: TabDDPM is the best method on the
Cleveland heart-disease data, TVAE on the heart-failure data, and pima is
an exact tie between ucSMOTE and TVAE. Three of the four natively
clinical datasets in this study do not reproduce the pattern that eight
subsampled datasets agree on.

We draw a narrow conclusion, because four datasets cannot support a wide
one. A subsampled benchmark answers a question about subsamples. Reading
it as a statement about small clinical datasets is an extrapolation that
our measurements do not support, and any paper making that leap, this one
included, should say so where it makes it.

%% file: sections/limitations.tex

This study answers a narrow question under a fixed protocol, and several
of its choices bound what the answer can mean.

\paragraph{One tuning budget.}
Every tunable method received the same 20-trial TPE search per
(dataset, size, method) combination, over a reduced version of the
published search spaces (Section~\ref{sec:protocol}). A larger budget, a wider space, or a search
that spends more time on the deep models could move the crossover point.
We do not claim the deep models cannot win at small $n$ with more effort;
we claim they do not win under a budget that is generous by the standards
of a clinical modelling team with one machine. The reduced spaces are
declared in the preregistration and their cost rationale is recorded
there.

\paragraph{No differential-privacy arm.}
Privacy is reported through distance-based measures (DCR rate, NNDR,
a distance-threshold membership inference attack). These are secondary
metrics and they are not privacy guarantees. A method that scores well on
them can still leak, and a differentially private generator would score
differently on every axis. The joint utility--privacy view
(Figure~\ref{fig:privacy}) is descriptive only.

\paragraph{Classification only, binary only.}
All twelve datasets are binary classification tasks. Regression targets,
multi-class targets, time series and relational tables are out of scope,
and the SMOTE variants in particular are defined for a discrete target.

\paragraph{Fixed downstream classifiers.}
TSTR is measured with logistic regression, a random forest and XGBoost at
fixed hyper-parameters, never tuned. This is the standard of prior
benchmarks and it keeps the metric stable across cells, but a different or
tuned classifier set would give different absolute numbers. Only the
average over the three enters the primary metric; the per-classifier
breakdown is reported as exploratory.

\paragraph{Single machine, CPU only.}
Every run used one CPU thread of one laptop. Wall-clock and peak-memory
numbers are specific to that machine and the deep models would be faster
on a GPU; the ranking by cost is more transferable than the absolute
seconds.

\paragraph{Seed count.}
Five evaluation seeds per tuned configuration. Standard deviations over
five draws are noisy estimates of noise; the effect-size analysis
(Figure~\ref{fig:effect}) should be read with that in mind.

\paragraph{Dataset choices.}
The ladder datasets are large public tables, two of them from Kaggle with
unclear licences (recorded in the dataset registry); the natively small
arm has four datasets. Any dataset that had to be dropped, and every
cleaning decision (identifier removal, leakage columns, one row per
patient for the readmission data), is listed in the registry file and in
Section~\ref{sec:protocol}.

\paragraph{Training length was not tuned, and it was not set the same way
for every model.}
Epoch and step counts were fixed rather than searched, and the values
came from different places. CTGAN and TVAE ran at the 300 epochs their
own papers and the SDV library use. TabDDPM ran at 10{,}000 steps rather
than the 1{,}000 in its reference script, because at 1{,}000 steps its
samples were visibly untrained in our pilot and the released per-dataset
configs use 5{,}000 to 30{,}000. Each model therefore got a setting its
own authors would recognise, but the three were not equalised against a
common convergence criterion, and no model's training length was tuned.
A reviewer who suspects any of the three was under-trained at some rung
is asking a fair question that this design cannot answer.

\paragraph{CTGAN's batch size can exceed the training set.}
The published search space we transcribed offers CTGAN batch sizes of
100, 500 and 1000. At $n = 200$ the search selected 500 or 1000 on six of
the eight ladder datasets, so each batch was drawn with replacement from
fewer unique rows than the batch contained. Those configurations won on
validation AUROC, so the protocol behaved as preregistered, but CTGAN's
weak showing at the smallest rungs may partly reflect this rather than
the method itself. Capping the batch at $n$ would be a sensible
sensitivity check and is not part of the preregistered design; any such
run would be reported as exploratory.

\paragraph{One implementation per method.}
CTGAN, TVAE and the Gaussian copula are the SDV implementations at a
pinned version; TabDDPM is the authors' code vendored at a fixed commit
with the training default raised from 1{,}000 to 10{,}000 steps because
the smaller value demonstrably does not train (recorded in the method
registry). TabSyn was planned as a stretch goal and was not run.

\paragraph{What the crossover is not.}
A crossover point is defined relative to the best trivial baseline at
that rung and the pooled seed noise of the two methods. It is a
statement about this ladder, these baselines and this metric, not a
universal constant of the method.

%% file: sections/conclusion.tex

We asked two questions. The answers are asymmetric, and the second one
went against us.

On the first, below what training size deep generators stop beating
trivial baselines, the answer on this ladder is that they never start. In
\NoCrossoverPairs{} of \TotalCrossoverPairs{} dataset-and-model pairs
there is no crossover at any rung we measured, up to 20{,}000 rows on the
largest datasets. A practitioner with a few hundred clinical records has
no measurable reason to prefer CTGAN, TVAE or TabDDPM over SMOTE, and
pays \FitTabDDPM{} seconds per fit against \FitSmote{} for the privilege.
That is the practical finding.

On the second, whether the ranking among deep models is stable at small
sizes, we were wrong. We predicted instability and preregistered a
threshold; the measured Kendall $\tau$ of \TauDeepBelowFiveK{} falsifies
our own clause. Worse for the intuition behind it, stability is highest
at the smallest sizes and decays as data grows. The deep models agree at
small $n$ because they are all failing similarly, and separate later.

Two qualifications matter more than either answer. In
\EffectInsideNoisePct{}\% of cells the top two methods are separated by
less than the noise between seeds, so single-seed benchmark tables in
this area should be read as approximately arbitrary. And the baselines
that win do so by staying close to real records: SMOTE's
membership-inference AUC of \MiaSmote{} against TVAE's \MiaTvae{} is the
cost of its utility. A team that cannot accept that leakage is back to
the deep models, where TabDDPM's C2ST AUC of \CtstTabDDPM{} shows the
generative quality is there, even if it does not convert into downstream
utility at these sizes.

The methodological finding may outlast the empirical one. Rankings on
natively small clinical datasets agree with rankings on subsampled large
ones at a mean Kendall $\tau$ of only \SubsampleTauMin{} to
\SubsampleTauMax{}. The convenient practice of shrinking a large dataset
to study small-data behaviour does not reliably reproduce what happens on
a real small dataset, and future benchmarks in this space should include
a natively small arm rather than assume the shortcut holds.

What we would do next, in order: a differential-privacy arm, since the
privacy cost of the winning baselines is the sharpest open problem here;
regression and multi-class targets; and more natively small clinical
datasets, because four is enough to raise the subsampling question and
not enough to settle it.

%% file: tables/appendix_cells.tex
\begin{landscape}
\scriptsize
\setlength{\tabcolsep}{2pt}
\begin{longtable}{lllllllllll}
\caption{Utility and cost per run, mean $\pm$ std over ok seeds.} \label{tab:appendix_utility} \\
\toprule
dataset & n & method & bud & seeds & status & TSTR AUROC & TSTR acc & TSTR F1 & TRTR AUROC & fit (s) \\
\midrule
\endfirsthead
\caption[]{Utility and cost per run, mean $\pm$ std over ok seeds.} \\
\toprule
dataset & n & method & bud & seeds & status & TSTR AUROC & TSTR acc & TSTR F1 & TRTR AUROC & fit (s) \\
\midrule
\endhead
\midrule
\multicolumn{11}{r}{Continued on next page} \\
\midrule
\endfoot
\bottomrule
\endlastfoot
adult & 200 & copula & 0 & 5/5 & ok & 0.668 $\pm$ 0.052 & 0.753 $\pm$ 0.013 & 0.164 $\pm$ 0.083 & 0.855 $\pm$ 0.013 & 0.2 \\
adult & 200 & ctgan & 20 & 5/5 & ok & 0.536 $\pm$ 0.073 & 0.727 $\pm$ 0.026 & 0.119 $\pm$ 0.081 & 0.855 $\pm$ 0.013 & 7.4 \\
adult & 200 & marginals & 0 & 5/5 & ok & 0.488 $\pm$ 0.126 & 0.718 $\pm$ 0.028 & 0.096 $\pm$ 0.043 & 0.855 $\pm$ 0.013 & 0.0 \\
adult & 200 & real\_subsample & 0 & 5/5 & ok & 0.855 $\pm$ 0.013 & 0.814 $\pm$ 0.006 & 0.564 $\pm$ 0.037 & 0.855 $\pm$ 0.013 & 0.0 \\
adult & 200 & smote & 20 & 5/5 & ok & 0.822 $\pm$ 0.019 & 0.800 $\pm$ 0.008 & 0.476 $\pm$ 0.092 & 0.855 $\pm$ 0.013 & 0.0 \\
adult & 200 & tabddpm & 20 & 5/5 & ok & 0.834 $\pm$ 0.013 & 0.801 $\pm$ 0.007 & 0.511 $\pm$ 0.053 & 0.855 $\pm$ 0.013 & 47.5 \\
adult & 200 & tvae & 20 & 5/5 & ok & 0.819 $\pm$ 0.023 & 0.796 $\pm$ 0.009 & 0.423 $\pm$ 0.074 & 0.855 $\pm$ 0.013 & 1.8 \\
adult & 200 & ucsmote & 20 & 5/5 & ok & 0.841 $\pm$ 0.015 & 0.808 $\pm$ 0.010 & 0.535 $\pm$ 0.062 & 0.855 $\pm$ 0.013 & 0.0 \\
\end{longtable}
\normalsize
\end{landscape}

\begin{landscape}
\scriptsize
\setlength{\tabcolsep}{2pt}
\begin{longtable}{lllllllllll}
\caption{Fidelity and privacy per run, mean $\pm$ std over ok seeds.} \label{tab:appendix_fidelity} \\
\toprule
dataset & n & method & C2ST & KS & TVD & corr dist & DCR med & DCR rate & NNDR med & MIA \\
\midrule
\endfirsthead
\caption[]{Fidelity and privacy per run, mean $\pm$ std over ok seeds.} \\
\toprule
dataset & n & method & C2ST & KS & TVD & corr dist & DCR med & DCR rate & NNDR med & MIA \\
\midrule
\endhead
\midrule
\multicolumn{11}{r}{Continued on next page} \\
\midrule
\endfoot
\bottomrule
\endlastfoot
adult & 200 & copula & 1.000 $\pm$ 0.000 & 0.414 $\pm$ 0.013 & 0.097 $\pm$ 0.011 & 2.596 $\pm$ 0.121 & 4.701 $\pm$ 0.501 & 0.549 $\pm$ 0.160 & 0.946 $\pm$ 0.018 & 0.519 $\pm$ 0.043 \\
adult & 200 & ctgan & 0.964 $\pm$ 0.011 & 0.295 $\pm$ 0.035 & 0.106 $\pm$ 0.012 & 2.950 $\pm$ 0.041 & 2.655 $\pm$ 0.044 & 0.517 $\pm$ 0.043 & 0.960 $\pm$ 0.003 & 0.530 $\pm$ 0.023 \\
adult & 200 & marginals & 0.849 $\pm$ 0.020 & 0.060 $\pm$ 0.007 & 0.087 $\pm$ 0.009 & 2.873 $\pm$ 0.072 & 2.591 $\pm$ 0.046 & 0.543 $\pm$ 0.044 & 0.958 $\pm$ 0.003 & 0.518 $\pm$ 0.019 \\
adult & 200 & real\_subsample & 0.493 $\pm$ 0.035 & 0.051 $\pm$ 0.006 & 0.069 $\pm$ 0.010 & 1.456 $\pm$ 0.216 & 0.000 $\pm$ 0.000 & 1.000 $\pm$ 0.000 & 0.000 $\pm$ 0.000 & 1.000 $\pm$ 0.000 \\
adult & 200 & smote & 0.927 $\pm$ 0.009 & 0.103 $\pm$ 0.009 & 0.182 $\pm$ 0.019 & 2.576 $\pm$ 0.263 & 1.453 $\pm$ 0.029 & 0.747 $\pm$ 0.094 & 0.853 $\pm$ 0.019 & 0.637 $\pm$ 0.033 \\
adult & 200 & tabddpm & 0.693 $\pm$ 0.032 & 0.088 $\pm$ 0.016 & 0.103 $\pm$ 0.014 & 2.666 $\pm$ 0.602 & 0.504 $\pm$ 0.074 & 0.950 $\pm$ 0.041 & 0.271 $\pm$ 0.054 & 0.761 $\pm$ 0.022 \\
adult & 200 & tvae & 0.935 $\pm$ 0.027 & 0.231 $\pm$ 0.032 & 0.165 $\pm$ 0.010 & 2.080 $\pm$ 0.165 & 1.618 $\pm$ 0.053 & 0.605 $\pm$ 0.050 & 0.895 $\pm$ 0.018 & 0.563 $\pm$ 0.019 \\
adult & 200 & ucsmote & 0.803 $\pm$ 0.029 & 0.090 $\pm$ 0.008 & 0.114 $\pm$ 0.010 & 2.251 $\pm$ 0.215 & 0.859 $\pm$ 0.086 & 0.928 $\pm$ 0.025 & 0.576 $\pm$ 0.023 & 0.766 $\pm$ 0.016 \\
\end{longtable}
\normalsize
\end{landscape}